\documentclass[11pt,a4paper]{article}
\usepackage[hyperref]{acl2020}
\usepackage{times}
\usepackage{latexsym}
\usepackage{graphicx}
\usepackage{multirow}
\usepackage{hyperref}
\graphicspath{ {./images/} }

\usepackage{microtype}

\aclfinalcopy 

\title{Conversation Learner -- A Machine Teaching Tool for Building
Dialog Managers for Task-Oriented Dialog Systems}

\author{\footnotemark[2]\enspace Swadheen Shukla\Thanks{  Equal contribution.} \and \footnotemark[2]\enspace Lars Liden\footnotemark[1] \and \footnotemark[2]\enspace Shahin Shayandeh\footnotemark[1] \and \footnotemark[3]\enspace Eslam Kamal \and  \footnotemark[2]\enspace Jinchao Li \\  \footnotemark[2]\enspace \textbf{Matt Mazzola} \and \footnotemark[2]\enspace \textbf{Thomas Park} \and \footnotemark[2]\enspace \textbf{Baolin Peng} \and \footnotemark[2]\enspace \textbf{Jianfeng Gao} \\ \\
  \footnotemark[2]\enspace Microsoft Research AI\\
  \footnotemark[3]\enspace Microsoft Corporation\\
  \small{\texttt{\{swads,laliden,shahins,eskam,jincli,mattm,thpar,bapeng,jfgao\}@microsoft.com}}}

\date{}

\begin{document}
\maketitle

\begin{abstract}
Traditionally, industry solutions for building a task-oriented dialog system have relied on helping dialog authors define rule-based dialog managers, represented as dialog flows. While dialog flows are intuitively interpretable and good for simple scenarios, they fall short of performance in terms of the flexibility needed to handle complex dialogs. On the other hand, purely machine-learned models can handle complex dialogs, but they are considered to be black boxes and require large amounts of training data. 
In this demonstration, we showcase \emph{Conversation Learner}, a machine teaching tool for building dialog managers.
It combines the best of both approaches by 
enabling dialog authors to create a dialog flow using familiar tools, 
converting the dialog flow into a parametric model (e.g., neural networks),
and allowing dialog authors to improve the dialog manager (i.e., the parametric model) over time by leveraging user-system dialog logs as training data through a machine teaching interface.
\end{abstract}

\section{Introduction}
\label{sec:intro}

The proliferation of messaging applications and hardware devices with personal assistants has spurred the imagination of many in the technology industry to create task-oriented dialog systems that help users complete a wide range of tasks through natural language conversations.
Tasks include customer support, IT helpdesk, information retrieval, appointment booking, etc. 
The wide variety of tasks has created the need for a flexible task-oriented dialog system development platform that can support many different use cases, while remaining simple for developers to use and maintain.

A task-oriented dialog system is typically built 
as a combination of three discrete systems, performing language understanding (for identifying user intent and extracting associated information), dialog management (for guiding users towards task completion), and language generation (for converting agent actions to natural-language system responses). The Dialog Manager (DM) contains two sub-systems: the Dialog State Tracker (DST) for keeping track of the current dialog state, and the Dialog Policy (DP) for determining the next action to be taken in a given dialog instance. The DP relies on the internal state provided by DST to select an action, which can be a response to the user, or some operation on the back-end database (DB). 
In this paper, we present a novel approach to building dialog managers (DMs).

\begin{figure}[ht]
\centering
\includegraphics[clip,width=\linewidth]{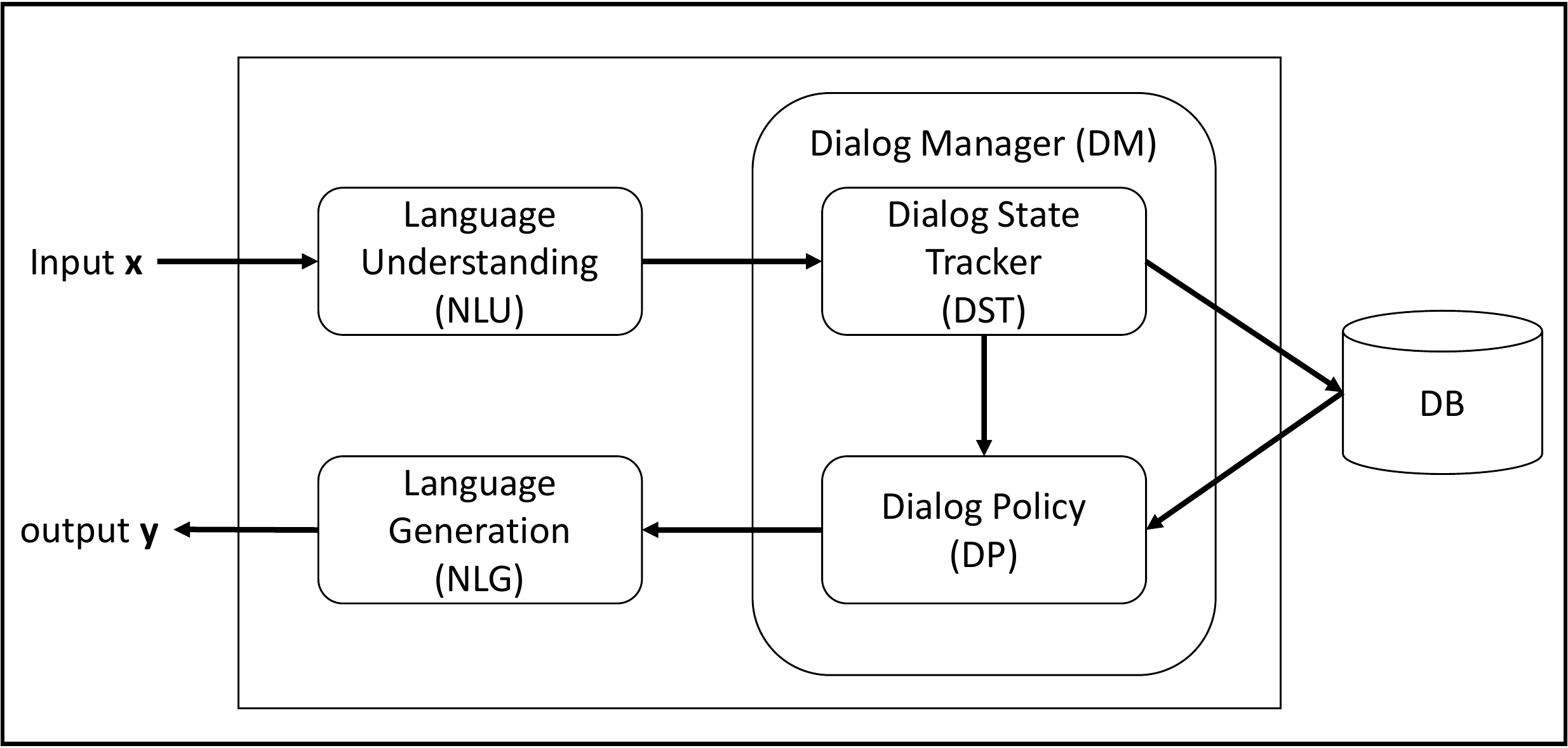}
\caption{An architecture for a task-oriented dialog system.}
\end{figure}

In a typical industrial implementation of a task-oriented dialog system, the DM is expressed as a \emph{dialog flow}, which is often a finite state machine, with nodes representing dialog activities (system actions) and edges representing conditions (dialog states that represent the previous user-system interactions). Since a dialog flow can be viewed as a set of \emph{rules} that specify the flow between dialog states, it may also be called a \emph{rule-based DM}.

There has been an increasing need for tools to help dialog authors\footnote{In this paper, ``author'' may refer to developers, business owners, or domain experts who define and maintain the conversational aspects of a task-oriented dialog system.} develop and maintain rule-based DMs. These tools are often implemented as drag-and-drop WYSIWYG tools that allow users to specify and visualize all the details of the dialog flow. They often have deep integration with popular Integrated Development Environments (IDEs) as editing frontends. Examples of rule-based or partially rule-based DMs include Microsoft's Power Virtual Agents (PVA) \footnote{\url{https://powervirtualagents.microsoft.com/}} and Bot Framework (BF) Composer \footnote{\url{https://github.com/microsoft/BotFramework-Composer}}, Google's Dialog Flow \footnote{\url{https://dialogflow.com/}}, IBM's Watson Assistant \footnote{\url{https://www.ibm.com/watson/}}, Facebook's Wit.ai \footnote{\url{https://wit.ai/}} and Amazon's Lex \footnote{\url{https://aws.amazon.com/lex/}}. It should be noted that most of these tools have some built-in machine-learned NLU capabilities, i.e. intent classification and entity detection, that can be leveraged to trigger different rule-based dialog flows, e.g. asking appropriate questions based on missing slots from the dialog state.

However, a rule-based DM suffers from two major problems. First, these systems can have difficulty handling complex dialogs. Second, updating a rule-based DM to handle unexpected user responses and off-track conversations is often difficult due to the rigid structure of the dialog flow, the long-tail (sparseness) of user-system dialogs, and the complexity in jumping to unrelated parts of the flow.

In end-to-end approaches proposed recently \cite{madotto-etal-2018-mem2seq,lei-etal-2018-sequicity}, the DM is implemented as a neural network model that is trained directly on text transcripts of dialogs. \citet{gao2019neural} presents a survey of recent approaches. One benefit provided by using a neural network model is that the network infers a latent representation of dialog state, eliminating the need for explicitly specifying dialog states. Neural-based DMs has been an area of active development for the research community as well as in industry; PyDial~\cite{ultes2017pydial}, ParlAI~\cite{parlai}, Plato~\cite{plato-2020}, Rasa~\cite{rasa}, DeepPavlov~\cite{deeppavlov}, and ConvLab~\cite{convlab} are a few examples. However, these machine-learned neural DMs are often viewed as black boxes from which dialog authors have difficulty interpreting why individual use cases succeed or fail. Further, these approaches often lack a general mechanism for accepting task-specific knowledge and constraints, thus requiring a large number of validated dialog transcripts for training. Collection and curation of this type of corpus is often infeasible.

This paper presents \href{https://www.microsoft.com/en-us/research/project/conversation-learner/}{Conversation Learner}, a machine \emph{teaching} tool for building DMs, which combines the strengths of both rule-based and machine-learned approaches. 
Conversation Learner is based on Hybrid Code Networks (HCNs) \cite{jason2017hcn} and the \emph{machine teaching} discipline \cite{machine-teaching-2017}.  
Conversation Learner allows dialog authors to (1) import a dialog flow developed using popular dialog composers, (2) convert the dialog flow to an HCN-based DM, (3) continuously improve the HCN-based DM by reviewing user-system dialog logs and providing updates via a machine teaching UI, and (4) convert the (revised) HCN-based DM back into a dialog flow for further editing and verification. 

Section 2 describes the architecture and main components of Conversation Learner. 
Section 3 demonstrates Conversation Learner features. 
Section 4 presents a case study of using Conversation Learner as the DM of a text-based customer support dialog system.

\section{Conversation Learner}

Development of any DM follows an iterative process of generation, testing, and revision.  
Conversation Learner follows a three-stage DM development process:
\begin{enumerate}
    \item Dialog authors develop a rule-based DM (dialog flow) using a dialog composer.
    \item The DM is imported into a HCN dialog system. Users (or human subjects recruited for system fine-tuning) interact with the system and generate user-system dialog logs.
    \item Dialog authors revise the DM by selecting representative failed dialogs from the logs and teaching the system to complete these dialogs successfully. Run regression testing. Return to step 2.
\end{enumerate}

\begin{figure}[ht]
\centering
\includegraphics[width=\linewidth,height=0.25\textheight,keepaspectratio]{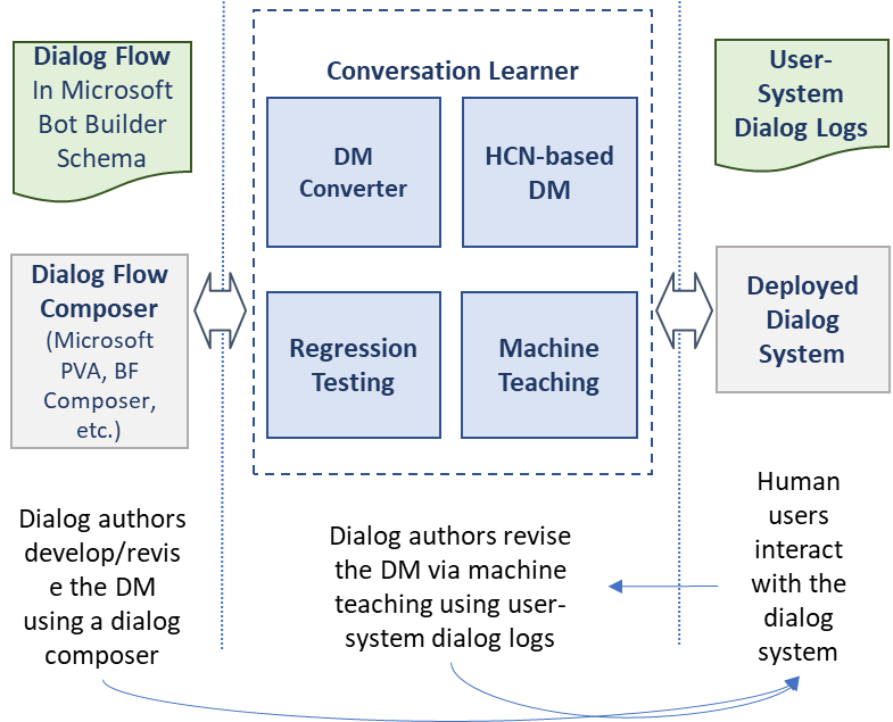}
\caption{The architecture of Conversation Learner (Top) and the development of DMs using Conversation Learner (Bottom).}
\label{fig:cl-architecture}
\end{figure}

This development process is illustrated in Figure~\ref{fig:cl-architecture} (Bottom).
The overall architecture of Conversation Learner is shown in Figure~\ref{fig:cl-architecture} (Top). 
It consists of four components: (1) a DM converter that converts a dialog flow between rule-based and HCN-based DM representations; (2) an HCN-based DM engine; (3) a machine teaching module that allows dialog authors to revise the HCN-based DM; and (4) an evaluation module that allows side-by-side comparison of the dialogs generated by different DMs. We describe each component in detail below.

\subsection{HCN-based DM}
\label{subsec-hcn-based-dm}

The Conversation Learner HCN consists of a set of task-specific action templates, an entity module, a set of action masks, and a Recurrent Neural Network (RNN). 
Each action template can be a textual communicative action, rich card, or an API call. The entity module detects entity mentions in user utterances, grounds the entity mentions (e.g., by mapping an entity mention to a specific row in a dataset), and performs entity substitution in a selected action template to produce a fully-formed action (e.g., by mapping the template ``the weather of [city]?'' to ``the weather of Seattle?''

Each action mask represents an ``if-then'' rule that determines the set of valid actions for some conditions (i.e., particular dialog states or user inputs).

The RNN maintains dialog states and selects system actions.
For each turn in a training dialog, a combination of features, including the user utterance embedding, its bag of words vector, and the set of extracted entities are concatenated to form a feature vector that is passed to the RNN, specifically a Long Short Term Memory (LSTM) network. The RNN computes a hidden state vector, which is retained for the next timestep. 
Next, a softmax activation layer is used to calculate a probability distribution over the available system action templates. 
An action mask is then applied, and the result is normalized to select the highest-probability action as the best response for the current turn.

The HCN can be trained on a collection of user-system dialogs. For each system response in a dialog, the action template is labeled. The training of HCN takes two steps. First, all unique action templates are imported into the HCN. Then, the RNN, which maps states to action templates, is optimized for minimizing the categorical cross-entropy on training data. More specifically, each dialog forms one minibatch, and updates of the RNN are done via non-truncated back-propagation through time.

Readers are referred to \citet{jason2017hcn} for a detailed description of HCN. It should be noted that CL leverages the same network architecture as HCN with enhancements and modifications to generate context features from training samples. 

\subsection{DM converter}
\label{subsec-dm-converter}

The DM converter converts a rule-based DM, developed using a dialog composer, to an HCN-based DM, which can then be improved via training dialogs and machine teaching. 

Given a dialog flow, the DM converter automatically generates a set of training dialogs that represent the dialog flow. 
This process is done by performing an exhaustive set of walks over the dialog flow and generating training dialog instances for each walk. 
Rules that determine transitions in the dialog flow are represented as action masks in the HCN. 
The HCN is trained on the generated training dialogs as described in Section \ref{subsec-hcn-based-dm}.

The DM converter can also convert a revised HCN-based DM back to a dialog flow by aggregating the individual training dialogs back into a graph for further editing and verification using a dialog composer.

\subsection{Machine Teaching}
\label{subsec-machine-teaching}

The HCN-based DM can be improved via machine teaching \cite{machine-teaching-2017}. ``Machine teaching'' is an active learning paradigm that focuses on leveraging the knowledge and expertise of domain experts as ``teachers''. This paradigm puts a strong emphasis on tools and techniques that enable teachers - particularly non-data scientists and non-machine-learning experts - to visualize data, find potential problems, and provide corrections or additional training inputs in order to improve the system's performance.

For Conversation Learner, we developed a UI for visualizing and editing logged user-system dialogs that had failed to complete their tasks successfully. The teacher does not need to revise the DM directly (e.g., via writing code or by modifying dialog structure in a hierarchical composer tool).
The teacher simply corrects cases where the dialog system responded poorly or incorrectly.

The teacher can make three types of corrections: (1) correct entity detection and grounding errors; (2) correct state-to-action mapping; or (3) create a new action template.

In cases where a large number of logged dialogs exists, we use active learning to provide a ranking of the candidate dialogs most likely to benefit from machine teaching intervention.

Our empirical tests show that machine teaching requires considerably fewer training samples than traditional machine learning approaches to improve system performance. 
We commonly observe significant improvements in DM performance by providing a dozen or fewer teaching examples.

There are three main reasons the HCN + machine teaching combination is so effective.
First, dialog authors are generally subject matter experts who can make well-informed decisions about which actions the DM should perform in individual situations.  
If the DM has to automatically learn an action policy from logs, a large corpus of data is required.
Second, the HCN allows dialog authors to explicitly encode domain-specific knowledge as action templates (bot activities or responses) and action masks (when a given response should be disallowed) without learning. 
Third, we can use intelligent filtering to select the most impactful failed dialogs for teachers to review.

\subsection{Regression Testing}

To effectively compare the performance of various dialog systems using different DMs, we developed a regression testing module. 
The module replays user utterances from transcripts of existing conversations against the DMs being tested; each DM then provides response action(s) for each turn. The module displays side-by-side comparisons of the resulting conversations from each DM, up to the point where the DM responses diverge.
Human judges then rate the conversations as ``left better'',``right better'' or ``same''. 

At the end of the rating session, a report is generated showing the performance of conversational flow amongst the DMs, as rated by the human judges.



\section{System Demonstration}

The system demonstration consists of the following steps:
\begin{enumerate}
    \item The dialog author creates a rule-based DM using a dialog composer tool. We showcase the usage of Microsoft Power Virtual Agent system for this step.
    \item The rule-based DM is exported to a common representation, e.g. Microsoft's `.dialog' schema\footnote{\url{https://github.com/microsoft/botbuilder-dotnet/blob/master/schemas/sdk.schema}}.
    \item The dialog flow is then imported into the Conversation Learner system. The import process involves traversing the dialog flow and creating dialog transcript instances for each traversal. 
    \item The generated dialog transcripts are run through both the rule-based system and Conversation Learner to demonstrate the improvement in dialog-handling flexibility provided by Conversation Learner's machine-learned DM. 
    \item For transcripts containing failures to complete the user's desired task, we demonstrate the machine teaching UI used to `correct' errors and generate new training examples for revising the DM.
\end{enumerate}

\section{Case Study}

To evaluate Conversation Learner's performance, we used dialog transcripts from Microsoft's text-based customer support system\footnote{These transcripts were scrubbed of any personally identifiable information.}. This customer support system uses a sophisticated rule-based DM to handle various customer support scenarios. It was created and is regularly updated by domain experts. We chose to use transcripts of actual user-system dialogs in order to measure the performance of the DMs in a realistic setting.

\begin{table}[ht]
\footnotesize
\begin{tabular}{|l|l|}
    \hline
    Turn & Dialog  \\ \hline
    \multirow{2}{*}{1}& usr: How to reduce font size \\&  agt: How can I help you with fonts today?\\ \hline
    \multirow{2}{*}{2}& usr: Adjust font size \\& \parbox{6cm}{agt: Would you like to change the font size in an app or the size of text on your screen?}\\ \hline
    \multirow{2}{*}{3}& usr: Change size of text on screen \\& \parbox{6cm}{agt: Change the size of text in Windows 10 using Display settings or use Magnifier. \textless Link to solution\textgreater} \\\hline
    \multirow{2}{*}{4}& agt: Did that solve your problem? \\& usr: Yes\\ \hline
    5 & agt: Great! Glad that helped. \\ \hline
\end{tabular}
\caption{A sample human-agent dialog from the customer support logs}
\label{tbl:sample-conversation}
\end{table}

\begin{figure}[ht]
\centering
\includegraphics[clip,width=\linewidth]{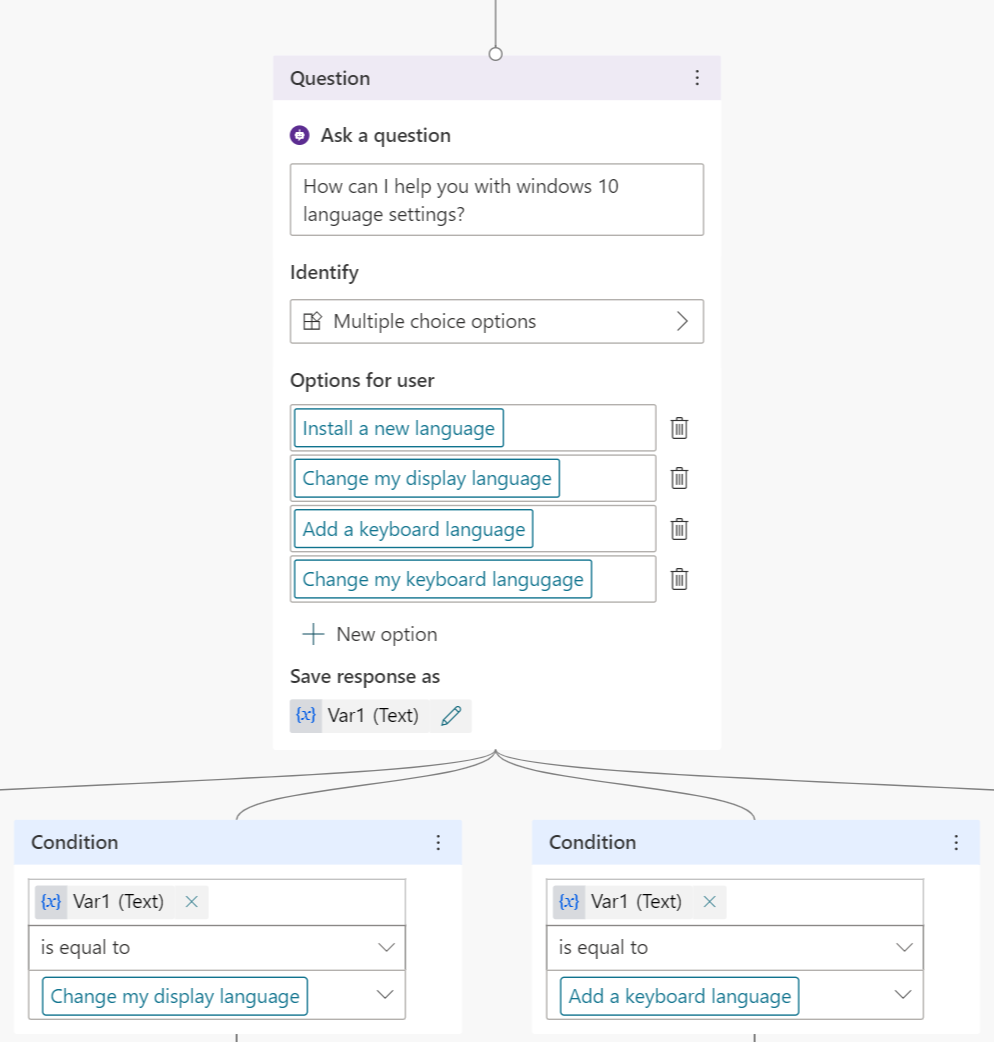}
\caption{An example of a rule-based dialog defined in the Microsoft Power Virtual Agents system}
\label{fig:sample-pva-dialog-flow}
\end{figure}

\begin{figure}[ht]
\centering
\includegraphics[clip,width=\linewidth]{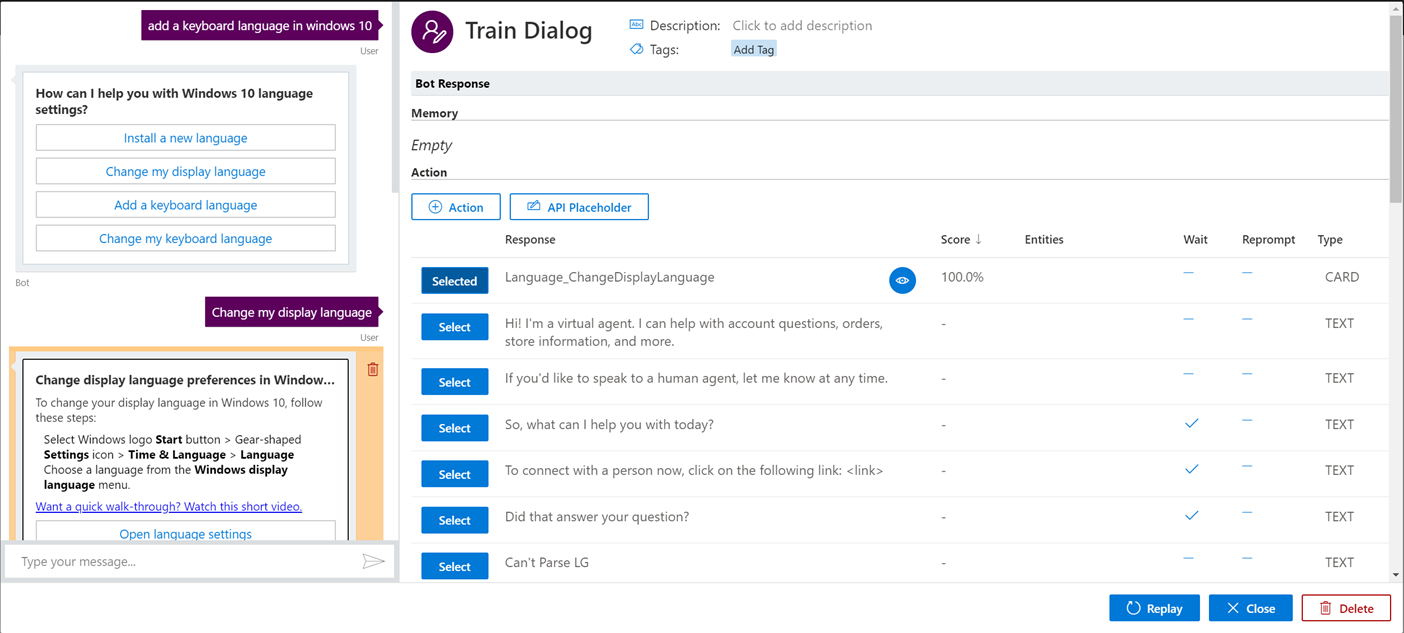}
\caption{An example train dialog generated by traversing the different paths of a dialog tree (left), and DM actions generated from the tree (right)}
\label{fig:cl-train-dialog}
\end{figure}

The customer support system consists of hundreds of ``topics'' out of which we selected three: ``language settings'', ``installing or changing fonts'', and ``sign in problems''. 
Table \ref{tbl:sample-conversation} shows an example of a user-system dialog for ``change font'' topic. Figure \ref{fig:sample-pva-dialog-flow} shows the representation of a specific topic's dialog flow in the Microsoft Power Virtual Agents \emph{dialog composer}. 
We exported the dialog flow graph of the support system from this system, then followed the process described in Section \ref{subsec-dm-converter} to train our machine-learned HCN-based DM. Figure \ref{fig:cl-train-dialog} shows an example of a generated train dialog.

After generating the HCN-based DM, we ran the set of dialog transcripts against both rule-based and HCN-based DMs. For the majority of conversations, users followed the expected flow, so the HCN-based DM produced the same results as the rule-based DM.  For those that differed, we used human judges to do a blind qualitative evaluation of the conversations and choose the conversation that provided the best task-completion result.

\begin{figure*}[ht]

\begin{minipage}{.45\linewidth}
\centering
  \includegraphics[width=\linewidth,height=0.3\textheight, keepaspectratio]{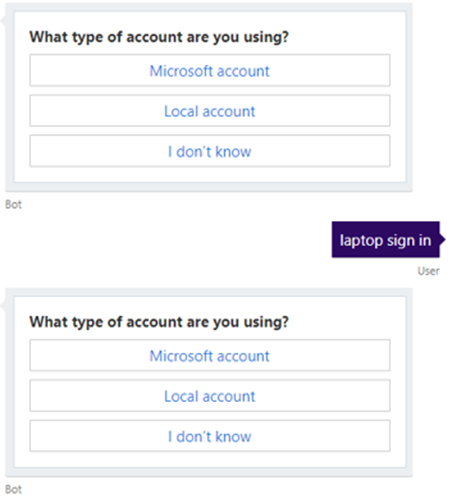}
  \captionof{figure}{A sample dialog from a rule-based system. Notice that the system just repeats its previous question (as a rule) since it did not understand the user reply.}
  \label{pva-sample-conv}
\end{minipage}
\hspace{.05\linewidth}
\begin{minipage}{.45\linewidth}
\centering
  \includegraphics[width=\linewidth,height=0.3\textheight, keepaspectratio]{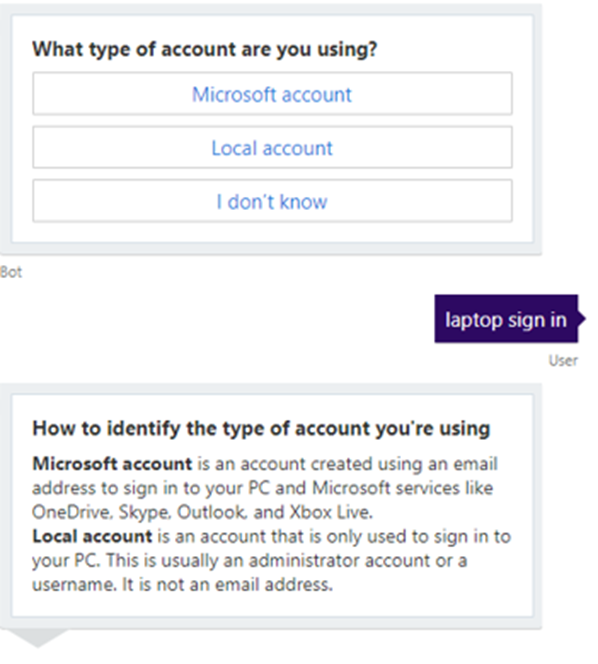}
  \captionof{figure}{Same dialog as Figure \ref{pva-sample-conv} in  Conversation Learner. Notice that the user response is accurately generalized to one of the available options when possible.}
  \label{cl-sample-conv}
\end{minipage}
\end{figure*}

\begin{table}[ht]
\begin{tabular}{|c|c|c|}
    \hline
    User Rating & \# of Convs. & \% \\ \hline
    CL is same &	2749 &	91.63\% \\ \hline
    CL is better &	136 &	4.53\% \\ \hline
    CL is worse &	115 &	3.83\% \\ \hline
    \multicolumn{2}{|l|}{Overall variation} & 0.7\% (better) \\ \hline
\end{tabular}
\caption{Initial results of human evaluation of 3000 dialogs against Conversation Learner (CL) and a rule-based dialog system.}
\label{init-result}
\end{table}

As shown in Table \ref{init-result}, the HCN-based DM provided better results for many transcripts, but there was an almost equal number of dialogs where the rule-based DM was rated better.  The rule-based DM may perform better in cases where specialized hard-coded logic was added to handle issues such as input normalization or rewriting.

An example dialog comparison is shown in Figures \ref{pva-sample-conv} and \ref{cl-sample-conv}. As seen in Figure \ref{cl-sample-conv}, the HCN-based DM handles cases where the user utterance does not match a known phrase, better than the rules-based system.  The HCN is also able to handle unexpected transitions between dialog nodes.



In a rule-based system, updating the DM to handle unexpected user responses and off-track conversations is much harder due to the rigid structure of the dialog flow graph, the long-tail nature of user-system dialog transcripts (a large number of sparse examples), and complexity in transitioning between unrelated parts of the dialog flow.

\begin{figure}[ht]
\centering
\includegraphics[width=\linewidth]{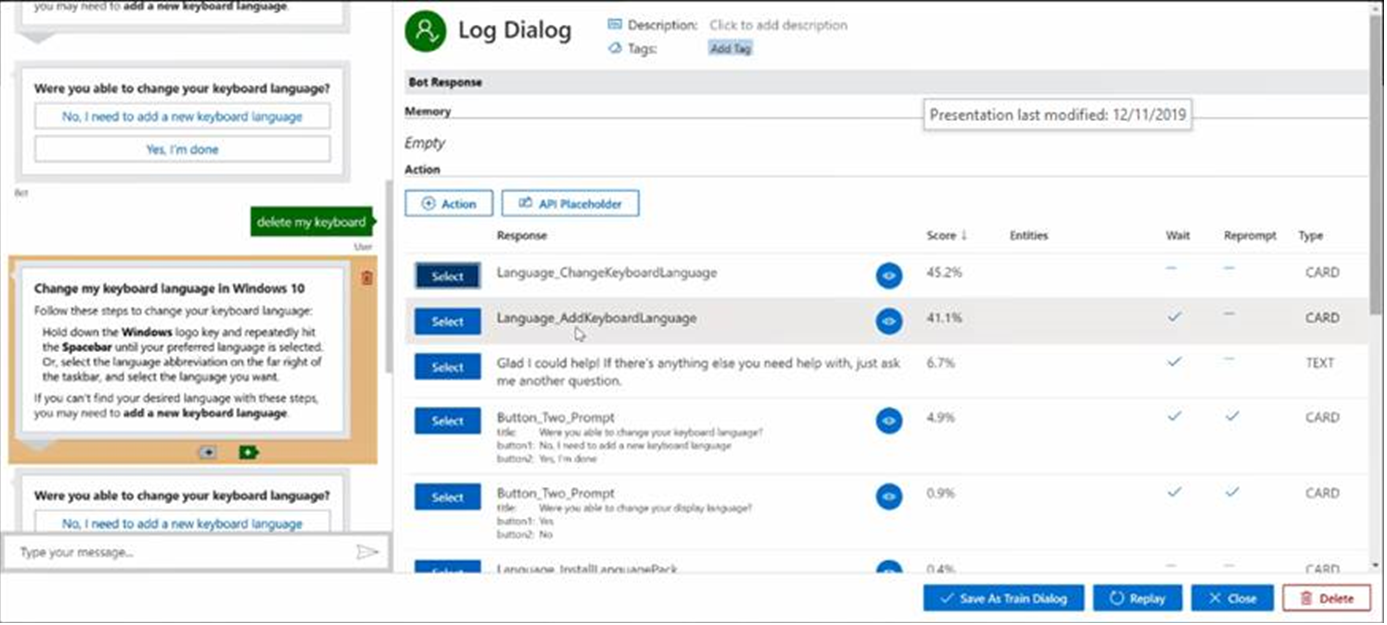}
\caption{Machine Teaching UI for \emph{correcting} dialogs to revise DM.}
\label{fig:machine-teach}
\end{figure}

Next, we demonstrate that the performance of the HCN DM can be substantially improved by adding just 3 to 5 teaching examples via the machine teaching UI presented in Section \ref{subsec-machine-teaching}. 
For this experiment, we chose dialog transcripts that contained common patterns of conversational problems, like users switching context, repeating themselves or asking follow-up questions, and corrected the dialog policy by creating or selecting the appropriate system action to resolve the problem. 
Once these additional examples are added, as illustrated in Figure \ref{fig:machine-teach}
the HCN-based DM's performance improvement over the rule-based DM nearly tripled, from 4.53\% to 13.8\%.

\begin{table}[ht]
\begin{tabular}{|c|c|c|}
    \hline
    User Rating & \# of Convs. & \% \\ \hline
    CL is same &	2562 &	85.4\% \\ \hline
    CL is better &	414 &	13.8\% \\ \hline
    CL is worse &	24 &	0.81\% \\ \hline
    \multicolumn{2}{|l|}{Overall variation} & 12.99\% (better) \\ \hline
\end{tabular}
\caption{Results of human evaluation of 3000 dialogs after improving Conversation Learner (CL) model with machine teaching.}
\label{teach-res}
\end{table}

As shown in Table \ref{teach-res}, minimal intervention from a dialog author by providing a small number of corrections to problematic user-system dialog logs can have a significant impact on the performance of the DM. As new users interact with the system and new transcripts are generated, the dialog author can continuously improve the HCN DM's performance by making corrections and adding new training data.

\section{Conclusion}
In this paper, we presented \emph{Conversation Learner}, a machine teaching tool for building dialog policy managers. 
We have shown that the CL HCN-based DM can be bootstrapped from a rule-based DM preserving the same behavior expected from the rule-based system. 
Using the CL machine teaching UI, the dialog author can provide corrections to the logged user-system dialogs and further improve the CL’s DM performance. 
We demonstrated this through a case study based on dialog transcripts from Microsoft’s text-based customer support system where the gains were approximately 13\%. 

We are planning to extend this work by looking into following problems: 1) Investigating effectiveness of different ranking algorithms for log correction recommendation, 
2) Optimizing number of training samples and action masks generated from the rule-based DM, and 3) Improving predictions of HCN-based DM by looking into alternative
network architectures.

\bibliography{acl2020}
\bibliographystyle{acl_natbib}

\end{document}